\documentclass[conference]{IEEEtran}

\newif\ifarxiv
\arxivtrue

\usepackage{cite}
\usepackage[utf8]{inputenc}
\usepackage[T1]{fontenc}
\usepackage{amsmath,amssymb,amsfonts}
\usepackage{booktabs}
\usepackage{multirow}
\usepackage{algorithmic}
\usepackage{graphicx}
\usepackage{textcomp}
\usepackage[dvipsnames]{xcolor}
\usepackage{xspace}
\usepackage{mathtools}
\usepackage[caption=false,farskip=0pt]{subfig}
\usepackage[acronym]{glossaries}
\usepackage{balance}
\usepackage[normalem]{ulem}
\usepackage{comment}
\usepackage[hyphens]{url}
\usepackage[colorlinks=true,allcolors=black, bookmarks=false]{hyperref}
\usepackage{arydshln}
\usepackage{diagbox}
\usepackage[normalem]{ulem}
\usepackage{gensymb} %
\usepackage[capitalise,noabbrev]{cleveref} %

\makeatletter
\DeclareRobustCommand\onedot{\futurelet\@let@token\@onedot}
\def\@onedot{\ifx\@let@token.\else.\null\fi\xspace}

\def\eg{e.g\onedot}

\def\ie{i.e\onedot}

\def\etal{\textit{et~al}\onedot}
\makeatother

\newcommand\major[1]{{#1}} %

\newacronymstyle{long-short-br}
{%
  \GlsUseAcrEntryDispStyle{long-short}%
}%
{%
  \GlsUseAcrStyleDefs{long-short}%
}
\setacronymstyle{long-short-br}

\def\BibTeX{{\rm B\kern-.05em{\sc i\kern-.025em b}\kern-.08em     T\kern-.1667em\lower.7ex\hbox{E}\kern-.125emX}}
\begin{document}

\title{Image-Based Automatic Dial Meter Reading\\in Unconstrained Scenarios}

\author{
\IEEEauthorblockN{Gabriel Salomon, Rayson Laroca, David Menotti}
\IEEEauthorblockA{Department of Informatics, Federal University of Paran\'a (UFPR), Curitiba, PR, Brazil \\
\, {\{\textit{gsaniceto, rblsantos, menotti}\}}\textit{@inf.ufpr.br}}
}
\maketitle

\ifarxiv
{\let\thefootnote\relax\footnote{This manuscript is a postprint of a paper accepted by \textit{Measurement}. See the final version on \textit{Science Direct} (DOI: \href{https://doi.org/10.1016/j.measurement.2022.112025}{\textcolor{blue}{10.1016/j.measurement.2022.112025}}).}}
\else
\fi

\newacronym{admr}{ADMR}{Automatic Dial Meter Reading}
\newacronym{amr}{AMR}{Automatic Meter Reading}
\newacronym{aneel}{ANEEL}{Brazilian Electricity Regulatory Agency}
\newacronym{ccw}{CCW}{counterclockwise}
\newacronym{cnn}{CNN}{Convolutional Neural Network}
\newacronym{copel}{Copel}{Energy Company of Paran\'a}
\newacronym{cw}{CW}{clockwise}
\newacronym{drr}{DRR}{dial recognition rate}
\newacronym{fcn}{FCN}{Fully Connected Network}
\newacronym{fcsrn}{FCSRN}{Fully Convolutional Sequence Recognition Network}
\newacronym{fps}{FPS}{frames per second}
\newacronym{gan}{GAN}{Generative Adversarial Network}
\newacronym{hct}{HCT}{Hough Circle Transform}
\newacronym{hlt}{HLT}{Hough Line Transform}
\newacronym{hog}{HOG}{Histogram of Oriented Gradients}
\newacronym{ht}{HT}{Hough Transform}
\newacronym{iou}{IoU}{Intersection over Union}
\newacronym{knn}{KNN}{K-Nearest Neighbors}
\newacronym{kwh}{kWh}{Kilowatt-hour}
\newacronym{lstm}{LSTM}{Long Short-Term Memory}
\newacronym{mae}{MAE}{Mean Absolute Error}
\newacronym{map}{mAP}{mean Average Precision}
\newacronym{mlp}{MLP}{Multilayer Perceptron}
\newacronym{mrr}{MRR}{meter recognition rate}
\newacronym{mser}{MSER}{Maximally Stable Extremal Regions}
\newacronym{mse}{MSE}{Mean Squared Error}
\newacronym{nms}{NMS}{Non-Maximum Suppression}
\newacronym{ocr}{OCR}{Optical Character Recognition}
\newacronym{rmse}{RMSE}{Root Mean Square Error}
\newacronym{rnn}{RNN}{Recurrent Neural Networks}
\newacronym{roi}{ROI}{Region of Interest}
\newacronym{rpn}{RPN}{Region Proposal Network}
\newacronym{sgd}{SGD}{Stochastic Gradient Descent}
\newacronym{sift}{SIFT}{Scale-Invariant Feature Transform}
\newacronym{ssd}{SSD}{Single Shot MultiBox Detector}
\newacronym{svm}{SVM}{Support Vector Machine}

\newcommand{\detnet}{Fast-YOLOv4-SmallObj\xspace}
\newcommand{\faster}{Faster R-CNN\xspace}
\newcommand{\mask}{Mask R-CNN\xspace}
\newcommand{\dataset}{UFPR-ADMR-v2\xspace}
\newcommand{\ufpradmr}{UFPR\nobreakdash-ADMR-v1\xspace}
\newcommand{\ufpramr}{UFPR-AMR\xspace}
\newcommand{\distance}{Levenshtein\xspace}

\newcommand{\crnn}{CRNN\xspace}
\newcommand{\grcnn}{GRCNN\xspace}
\newcommand{\rare}{RARE\xspace}
\newcommand{\rosetta}{Rosetta\xspace}
\newcommand{\rtwoam}{R\textsuperscript{2}AM\xspace}
\newcommand{\starnet}{STAR-Net\xspace}
\newcommand{\trba}{TRBA\xspace}
\newcommand{\vitstr}{ViTSTR\xspace}
\newcommand{\vitstrtiny}{ViTSTR-Tiny\xspace}
\newcommand{\vitstrsmall}{ViTSTR-Small\xspace}
\newcommand{\vitstrbase}{ViTSTR-Base\xspace}

\newcommand{\multitaskgabriel}{Multi-task\xspace}

\newcommand{\crnet}{CR-NET\xspace}
\newcommand{\fastocr}{Fast-OCR\xspace}

\newcommand{\supplementary}{https://github.com/guesalomon/ufpr-admr-v2-dataset/}

\newcommand{\nummodels}{18\xspace}
\ifarxiv
\vspace{-3.45mm}
\else
\fi
\begin{abstract}
The replacement of analog meters with smart meters is costly, laborious, and far from complete in developing countries.
The \gls*{copel} (Brazil) performs more than 4 million meter readings (almost entirely of non-smart devices) per month, and we estimate that 850 thousand of them are from dial meters.
Therefore, an image-based automatic reading system can reduce human errors, create a proof of reading, and enable the customers to perform the reading themselves through a mobile application.
We propose novel approaches for \gls*{admr} and introduce a new dataset for \gls*{admr} in unconstrained scenarios, called \dataset.
Our best-performing method combines YOLOv4 with a novel regression approach (AngReg), and explores several post-processing techniques.
Compared to previous works, it decreased the \gls*{mae} from 1,343 to 129 and achieved a \gls*{mrr} of 98.9\% –-~with an error tolerance of 1~\gls*{kwh}.

\end{abstract}

\begin{IEEEkeywords}
Automatic Meter Reading, Dial Meters, Pointer-type Meters, Deep Learning, Public Dataset
\end{IEEEkeywords}

\section{Introduction}
\label{sec:introduction}

\glsresetall

Meter reading is a laborious task and usually requires an employee to visit each meter periodically in order to measure the gas, water and energy consumption of consumers around the world~\cite{vanetti2013gas,gallo2015robust,li2019light}.
Although the world currently has the technology to perform readings using devices such as smart meters~\cite{kabalci2016survey}, there are still millions of old (analog and electronic) meter devices in operation, even in developed countries~\cite{salomon2020deep}.

\subsection{\major{Research Motivation}}

Replacing all of the deployed meters is an expensive and time-consuming procedure, still far from~complete~\cite{salomon2020deep,b0}.

There are two  main  categories  of energy meters: (i)~analog (i.e., with cyclometer and dial displays) and (ii)~digital (i.e., with electronic display, and smart meters)~\cite{salomon2020deep}.
Most works in the literature focus on digit-based meters (cyclometer and digital ones)~\cite{laroca2021towards, naim2021fully, azeem2020robust}.
However, the \gls*{copel} (presented in Section~\ref{sec:dataset}) performs more than $4$ million energy meter readings per month, of which we estimate that $850$ thousand are from dial meters (based on statistical sampling of $1$ million images provided by Copel). 
This means that there is still a significant percentage of these meters in use.
Regarding the automatic reading of pointer-based meters (dials and gauges), most works focus on industrial gauges such as pressure meters, ammeters, and voltmeters~\cite{he2019, liu2019,huang2019,cai2020pointer}. 

\gls*{admr} stands for the automatic reading of dial-based (or pointer-based) meters.

\begin{figure}[!htb]
  \centering
  \includegraphics[width=0.97\linewidth]{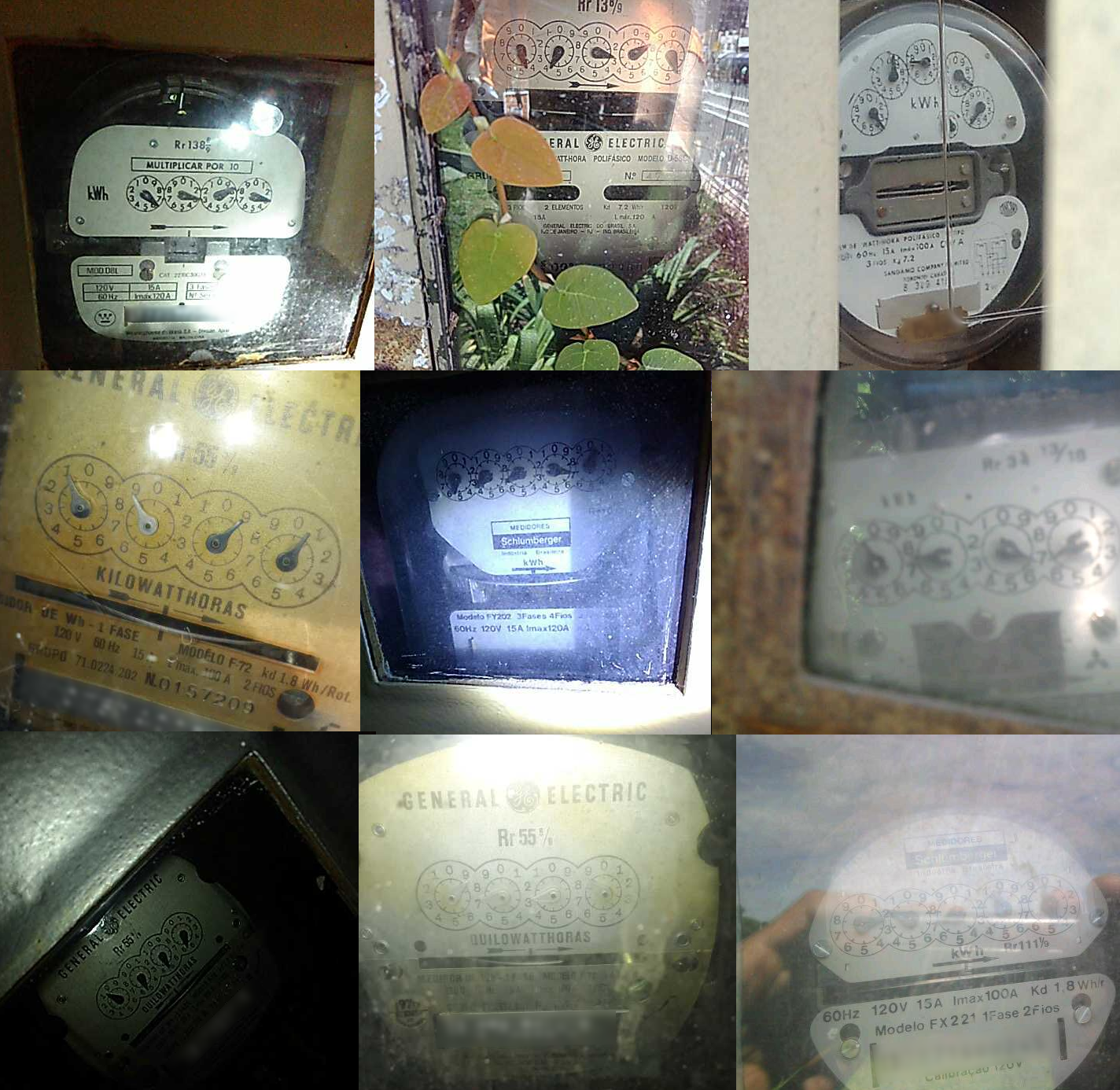}
  \vspace{-1.5mm}
   
    \caption{Representative samples from the proposed \dataset dataset. All images were taken in unconstrained scenarios, with significant variations in lighting, distance, contrast, noise, rotation and camera.
    }
    \label{fig:illustration-dataset}
\end{figure}

In this work, we tackle the issue of reading multi-dial meters using images (see Fig.~\ref{fig:illustration-dataset}) since it requires no change to the meters, being a faster, easier and cheaper solution to deploy than smart meters, which require the deployment of a data network (wired or wireless) and the replacement of the~meter.

\subsection{\major{Literature Review}}

Although there are many published works focused on image-based \gls{admr} systems, most of the proposed methods are designed and evaluated exclusively in well-controlled environments~\cite{howells2021real, fang2019, he2019}, \ie, on images with no challenging factors that are found on a daily basis by readers such as inaccessible meters (leading to oblique views), variable lighting conditions, distant meters, reflections, occlusion, blur, and wear caused by weather, rust and dust.
This is in stark contrast to related research areas, where in recent years the research focus shifted to unconstrained~scenarios~\cite{laroca2021towards}.
In addition, the majority of works do not provide a public dataset in order to enable reproducibility and fair comparison between approaches~\cite{howells2021real, fang2019, he2019}.
There is also a lack of state-of-the-art methods being explored, with a lot of handcrafted methods being used~\cite{jiale2011, zhang2016, zheng2017, vega2013}.
This is one of the rare works to employ deep learning models for the entire \gls{admr} pipeline.

\subsection{\major{Necessity of the Research}}
Considering electricity losses~\cite{da2019image}, according to the 2019 annual report of the \gls*{aneel}~\cite{aneel_report}, approximately $46\%$ of the company's non-technical electricity losses are caused by issues such as theft, tampering with meters, non-working meters, reading and billing errors. 
It is unclear; however, the occurrence ratio for each of the aforementioned events.
In absolute values, non-technical losses represent $33.3$~TWh per year ($6.6\%$ of the generated~electricity).

In this regard, a robust image-based \gls*{amr} system can speed up the reading process, reduce human errors, create proof of reading, and enable the customers to perform the reading themselves through a mobile application, eliminating the need for employees of the service company traveling around to perform local meter reading at each
consumer~unit.

\subsection{\major{Novelty and Main Contributions}}

In summary, the main contributions of this work are:
\begin{itemize}
    \item A public dataset for \gls*{admr}, called \dataset (see Fig.~\ref{fig:illustration-dataset}), that includes $5{,}000$ fully-annotated images of dial meters (totaling $22{,}410$ annotated dials) acquired in unconstrained scenarios by the service company’s employees themselves\footnote{The \dataset dataset is publicly available to the research community at \url{\supplementary}. Access is granted \textbf{upon request}, i.e., interested parties must register by filling out a registration form and agreeing to the dataset's terms of use.}.
    The proposed dataset contains a well-defined evaluation protocol to assist the development of new approaches toward this problem as well as to enable a quantitative and qualitative comparison among published works. To the best of our knowledge, this is the largest publicly available dataset in the \gls*{admr}~literature;
    \item A novel approach toward image-based \gls*{admr} that combines a detection model with our proposed regression approach (AngReg) in order to enable error correction for the most significant dials, thus decreasing the \gls*{mae} considerably. 
    We remark that most of the reading errors made by this approach occurred in challenging cases, where even humans can make mistakes due to artifacts in the counter~region;
    \item The most extensive comparative assessment in the \gls*{admr} literature, with $20$ approaches ($\nummodels$ models) being adjusted and evaluated under exactly the same conditions.
    We are not aware of any work in the literature where so many methods were evaluated in the experiments.
\end{itemize}

A preliminary version of this work was published at the 2020 International Joint Conference on Neural Networks~(IJCNN)~\cite{salomon2020deep}.
\major{This work differs from that in several aspects.
Here, we present and assess novel approaches for \gls*{admr}, such as regression and segmentation-free methods.
We also evaluate more detection-based methods, such as YOLOv4~\cite{bochkovskiy2020yolov4} and the end-to-end approach proposed in~\cite{laroca2021towards}.
In addition to expanding the \ufpradmr dataset considerably, with new images (from $2{,}000$ to $5{,}000$) and refined annotations, we provide solutions for each of the primary causes of reading errors found in~\cite{salomon2020deep}.
In this way, we attain much better results than those reported in our previous work, especially considering that $80$\% of the reading errors made by the best performing approach --~YOLOv4 (R) + (X)~-- occurred on the least significant dial.
In fact, $85$\% of all errors have \gls*{mae}~$=1$.
With an error tolerance of $1$ \gls*{kwh}, our final system achieves an impressive \gls*{mrr} of~$98.9$\%.}

\subsection{\major{Organization of the Paper}}
The remainder of this paper is organized as follows.
We review related works in Section~\ref{sec:related}.
The \dataset dataset is introduced in Section~\ref{sec:dataset}.
The proposed approaches are presented in Section~\ref{sec:approaches}.
In Section~\ref{sec:results}, we report the experiments performed and discuss the results obtained.
Finally, conclusions and future work are given in Section~\ref{sec:conclusions}.
\section{Related Work}
\label{sec:related}

Most works in the \gls*{amr} literature focus on digit-based meters (cyclometers and digital meters).
The literature for \gls*{admr} is more scarce.
Most of the works on pointer recognition focus on industrial gauges, with handcrafted methods and well-controlled environments regarding lighting, weather, and image quality; some of them used only images from the same distance and angle~\cite{huang2019, zheng2017, jiale2011, tang2015}. 
Recently, deep learning approaches have gained traction in the \gls*{amr}~\cite{laroca2021towards,laroca2019convolutional,tsai2019,yang2019} and \gls*{admr}~\cite{salomon2020deep, fang2019, he2019, hou2021automatic, cai2020pointer} contexts.
In this section, we review some relevant works on \gls*{admr}.
We also briefly describe some digit-based approaches that achieved promising results.
Lastly, we conclude this section with final~remarks.

Several works explored handcrafted features, such as \gls*{ht}~\cite{jiale2011, zhang2016, zheng2017, zheng2016robust}, to locate the dials or the pointers.
The steps usually include image binarization on the preprocessing stage, \gls*{hct} for dial detection, and pointer angle detection using \gls*{hlt} or similar~methods.
These approaches generally work well in constrained environments but may not be suitable for real-world scenarios (i.e., with uneven lighting and the presence of~noise)~\cite{luz2018deep,laroca2021towards}. 

\mask was proposed for reading meters based on pointers in~\cite{fang2019, he2019, zuo2020robust}.
Fang~\etal~\cite{fang2019} used it to find reference key points on both the pointer and the gauge scale marks, whereas He~\etal~\cite{he2019} and Zuo~\etal~\cite{zuo2020robust} focused on segmenting the dial and the pointer.
In these three works, the angle between the pointer and the respective dial was explored to retrieve the meter reading.
Although promising results were achieved in all of them, only private datasets were used in the~experiments.

Hou~\etal~\cite{hou2021automatic} explored the use of a lightweight CNN model to read cropped industrial gauges (the detection of the region of interest was not addressed in their work).
On the other hand, region-based \glspl*{fcn} were explored for meter detection by Huang et al.~\cite{huang2019}, which applied handcrafted methods in the remainder of the~pipeline.

Liu~\etal~\cite{liu2019} explored four  \gls*{cnn}-based object detectors (i.e., Fast R-CNN, \faster, \gls*{ssd}, and YOLO) for meter detection.
High \gls*{map} values were achieved for this task, however, no deep learning-based approach was evaluated at the recognition stage.

In~\cite{sowah2021intelligent}, Sowah~\etal extracted high-level features using a cascade of contour filters and performed the final recognition using a regression model and \gls*{knn}. 
Although only handcrafted features were explored, it is noteworthy that this is one of the few works where experiments were carried out with different models of~gauges.

Cai~\etal~\cite{cai2020pointer} and Howells~\etal\cite{howells2021real} tackled the lack of publicly available gauge datasets by introducing approaches to generate synthetic data. Cai~\etal~\cite{cai2020pointer} first generated synthetic samples by varying the center and angle of the pointer and then trained a \gls*{cnn} model to perform regression of the pointed values.
Howells~\etal\cite{howells2021real} leveraged synthetic 3D-rendered gauge images to train two \gls*{cnn} detectors, one for gauge detection and one for gauge reading. Their system corrects each gauge's perspective and identifies the scale marks to predict the reading correctly.
Although their system outperformed other methods for gauge reading, it was evaluated exclusively on images with the gauge well centered and occupying a large portion of the image, \ie, the images used in their experiments are not as unconstrained as those collected in the field by the service company’s~employees.

The following approaches focused on multi-dial meters, more specifically energy meters. Vega~\etal~\cite{vega2013} exploited handcrafted features to perform dial recognition in energy meters. Methods such as binarization and line intersection were used to extract features and the \gls*{sift} was employed to detect the counter region. 
This method was evaluated on a private dataset containing only $141$ well-controlled~images.

In~\cite{tang2015}, Tang~\etal proposed a complete framework for dial energy meter reading based on binarization, line intersection, and morphological operations.
As in~\cite{vega2013}, the images used in the experiments were obtained in a well-controlled environment and belong to a private~dataset.

In our previous work~\cite{salomon2020deep}, we compared several deep learning-based methods for \gls*{admr}.
More specifically, we fine-tuned object detectors (e.g., \faster~\cite{ren2015faster} and YOLOv3~\cite{redmon2018yolov3}) for dial recognition directly on the input images, \ie, without the need for \gls*{roi} detection or~segmentation.
Based on our experimental evaluation, performed on images from the \ufpradmr dataset, we highlighted the trade-off between precision and speed achieved by~YOLOv3 since it reached recognition rates competitive to \faster while being able to process three times~faster.

There are some digit-based \gls*{amr} research worth mentioning~\cite{gomez2018cutting,laroca2019convolutional,laroca2021towards}. 
Gómez~\etal~\cite{gomez2018cutting} introduced a CNN-based segmentation-free approach to perform digit-based meter reading, without the need for counter region detection. They achieved promising results in a private dataset. Nonetheless, the authors only evaluated their method against traditional algorithms that rely on handcrafted features, which are not robust to unconstrained scenarios~\cite{laroca2021towards}.
Laroca~\etal~\cite{laroca2019convolutional} developed a two-stage approach to tackle \gls*{amr}.
Fast-YOLOv2~\cite{redmon2017yolo9000} was employed for counter detection and three distinct \glspl*{cnn} were evaluated in the recognition stage.
As their two-stage approach failed in some cases where the counter region was tilted, in \cite{laroca2021towards} authors from the same research group designed a multi-task network to detect the four corners of the counter region to rectify it before the recognition~stage.
In their experiments, the improved system outperformed all baselines in terms of recognition rate while still being quite~efficient.

\subsection{Remarks}

Most of the approaches  were designed specifically for reading single-dial meters, which are also known as gauges. Even though the problems are similar, there are three fundamental differences between them: (i)~a small error in a multi-dial meter may result in a substantial reading error (in case the error~(s) occur in the most significant dials); (ii)~there are intersections between dials in several meters, which may interfere with the reading process and segmentation-based methods; (iii)~in gauges, the rightmost dials cannot be used for correction (as gauges usually only have one dial). More details on this type of correction are provided in Section~\ref{sec:results}. 

Most of the works performed experiments exclusively on private datasets.
Although there are a few publicly available datasets for \gls*{amr} (e.g.,~\cite{yang2019,nodari2011multi, laroca2019convolutional, naim2021fully, laroca2021towards, iqbal2021automated}),  none of them have images containing pointer-type meters, only digit-based ones. 
As far as we know, the only dataset with images of multi-dial meters currently available in the literature is \ufpradmr, which we introduced in the preliminary version of this work~\cite{salomon2020deep}. Besides, the majority of works deal with well-controlled and constrained scenarios, which are immensely simpler than real-world~scenarios.

\major{As mentioned in Section~\ref{sec:introduction}, there are several differences between this work and~\cite{salomon2020deep}.
For example, this work presents and assesses novel approaches for \gls*{admr}, such as regression and segmentation-free methods.
We emphasize that, to the best of our knowledge, no previous pointer-based works explored \gls*{lstm}-based approaches to tackle the reading process as a sequence problem.
We consider this a promising approach that can leverage the information of the previous dial to predict the current one, instead of reading each dial as an independent~instance.}

\major{
We also evaluate more detection-based methods, such as YOLOv4~\cite{bochkovskiy2020yolov4} and the end-to-end approach proposed in~\cite{laroca2021towards}, and expand the \ufpradmr dataset considerably (from $2{,}000$ to $5{,}000$ images).
Lastly, we tackle the primary sources of errors found in~\cite{salomon2020deep}, providing solutions to them (i.e., we explored data augmentation and post-processing techniques to reduce errors), thus increasing the \gls*{mrr} from $75.25$\% to $92.50$\% and reducing the \gls*{mae} from $1,343.29$ to~$187$.}

\major{
We reinforce that most works in the \gls*{admr} literature do not provide error analyses.
There are also not many post-processing methods available in the literature, that is, some works perform angle corrections, but none of them proposed a more robust multi-dial correction~method.}
\section{The \dataset Dataset}
\label{sec:dataset}

In our previous work~\cite{salomon2020deep}, we introduced the \ufpradmr dataset, which contains $2{,}000$ images acquired in the field by the employees of the \acrfull*{copel}, a company of the Brazilian electricity sector that serves more than $4$~million consuming units per month~\cite{copel, laroca2019convolutional}.

In this work, we introduce an expanded version of that dataset, called \dataset, with $3{,}000$ more images --~totaling $5{,}000$ images (with $22{,}410$ dials)~-- and new annotations such as the approximated pointed value labels (with one decimal place precision) and segmentation masks for the~dials.
To the best of our knowledge, this is the largest and most complete publicly available dataset in the \gls*{admr}~literature.
Some representative samples are shown in Fig.~\ref{fig:illustration-dataset} (in Section~\ref{sec:introduction}). 

All $5{,}000$ images were obtained on-site by \gls*{copel} employees as proof of reading.
For this task, cell phones of various brands were used, most of them with low-end cameras.
The images were compressed and resized for storage, thus reducing their overall quality.
In this sense, the images have a resolution of $640\times480$ or $480\times640$ pixels, with an average size of $60$~kB.

\begin{figure}[!htb]
    \centering

    \subfloat[blur]{
        \includegraphics[width=.3\linewidth]{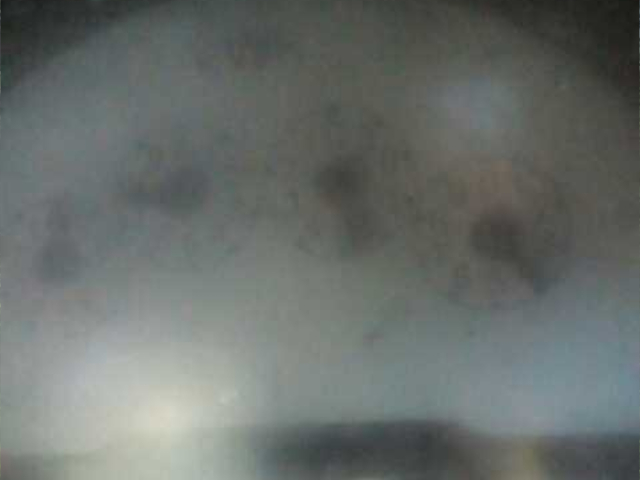}
        
    }
    \subfloat[weathering]{
        \includegraphics[width=.3\linewidth]{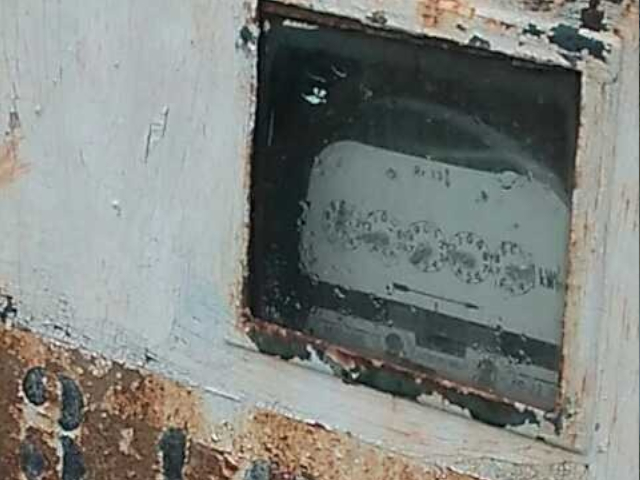}
    }
    \subfloat[distant capture]{
        \includegraphics[width=.3\linewidth]{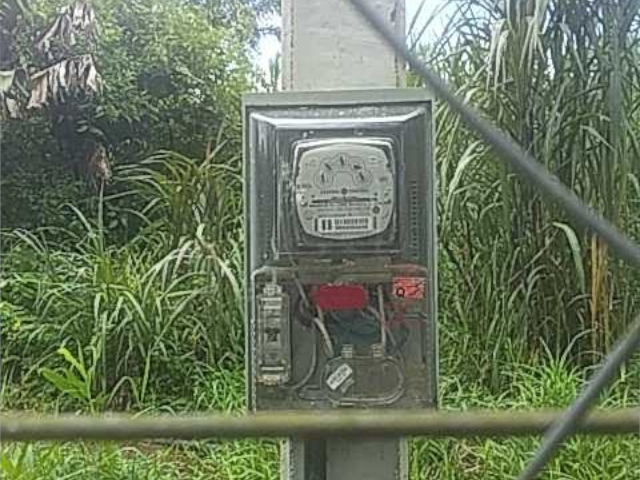}
    }

    \vspace{2.5mm}

    \subfloat[occlusion]{
        \includegraphics[width=.3\linewidth]{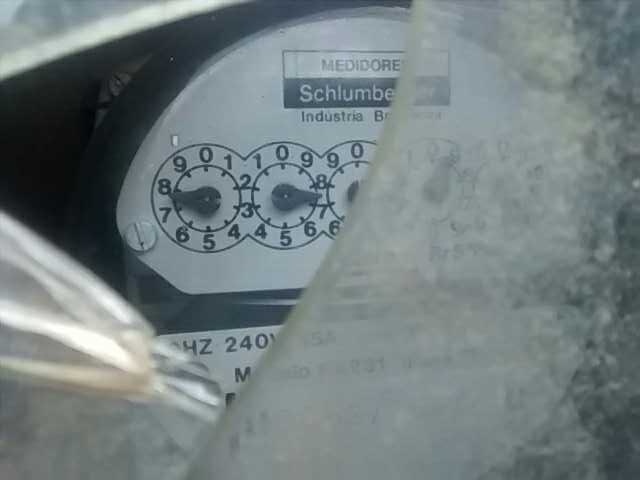}
    }
    \subfloat[dirt]{
        \includegraphics[width=.3\linewidth]{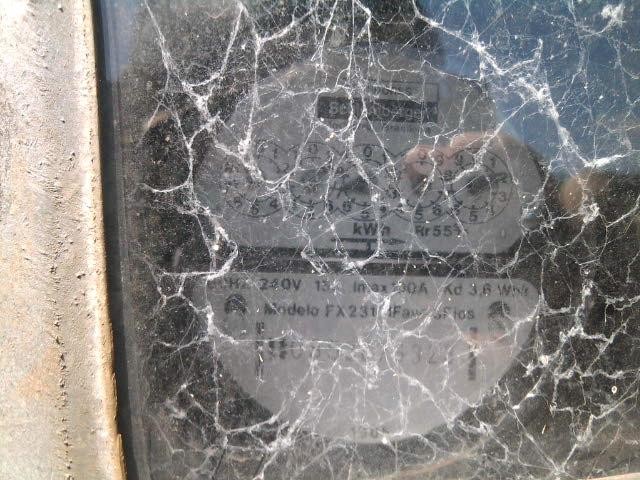}
    }
    \subfloat[reflections]{
        \includegraphics[width=.3\linewidth]{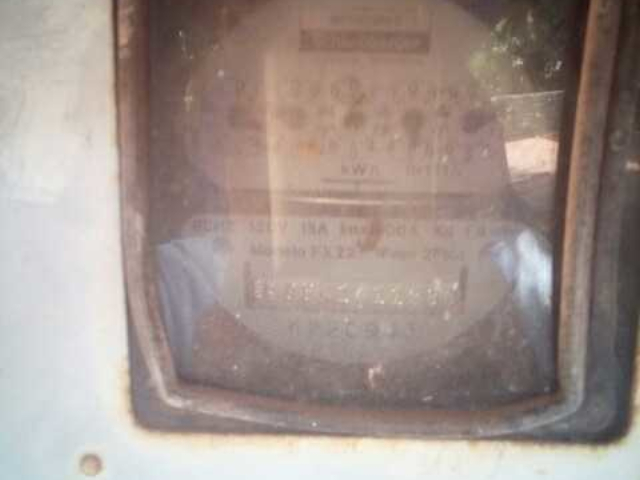}
    }
    
    \caption{Some of the challenging scenarios in the \dataset dataset. 
    For evaluation purposes, we only selected images where it is possible for a human to recognize the correct meter reading.
    }
    \label{fig:challenging-scenarios}
\end{figure}

There are many challenges a robust system must be able to handle in order to correctly perform meter readings in real-world scenarios.
As illustrated in Fig.~\ref{fig:challenging-scenarios}, these challenges include inaccessible meters (resulting in oblique views), variable lighting conditions, distant meters, reflections, occlusion, blur, and the wear caused by weather, rust and dust.
Furthermore, dial meters are subject to mechanical failure such as misalignment, non-working pointers, among others, which undermine the performance of an \gls*{admr}~system.

\begin{figure}[!htb]
    \centering
    
    \subfloat[$4$-dial meter]{
        \includegraphics[width=0.76\linewidth]{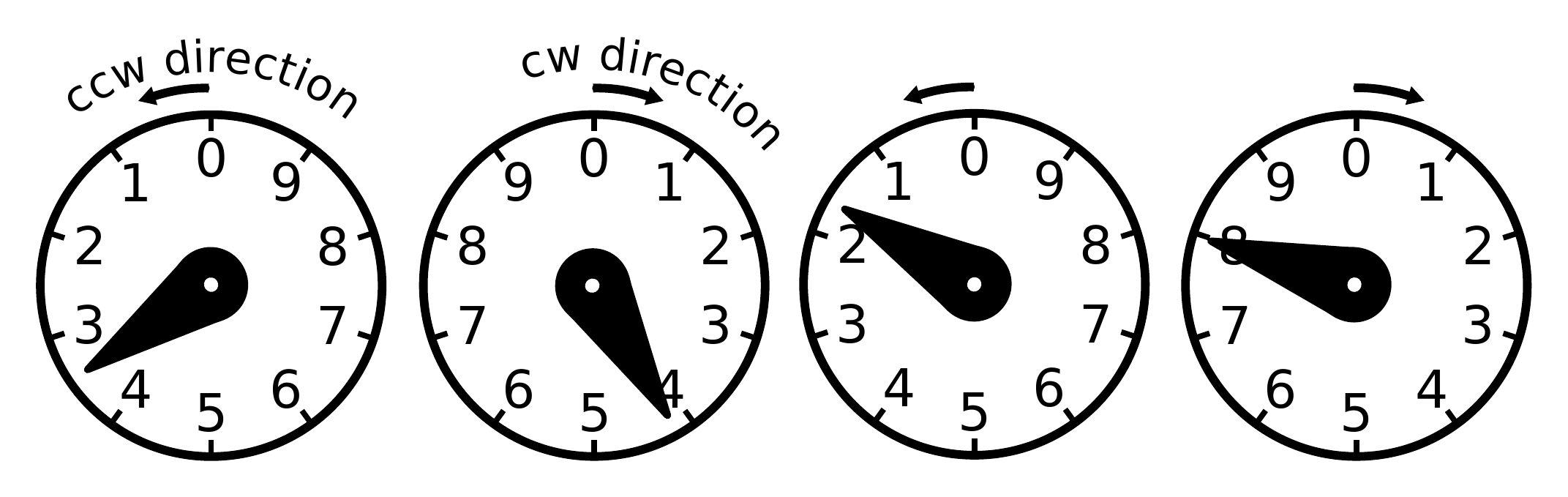}
    }
    
    \vspace{0.25mm}
    
    \subfloat[$5$-dial meter]{
        \includegraphics[width=\linewidth]{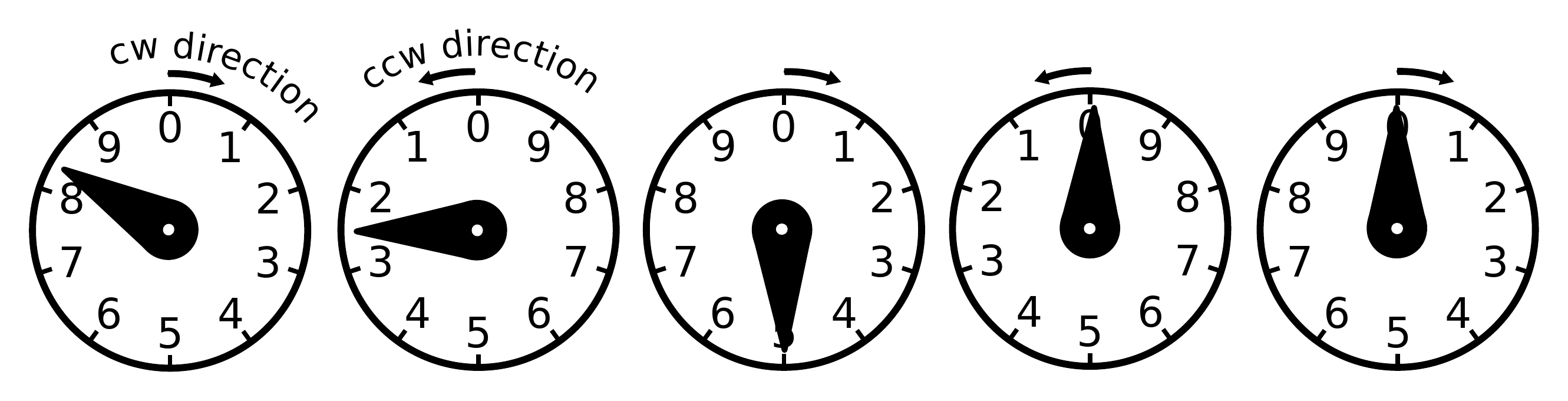}
    }
     
    \caption{Illustration of the energy meter dials present in the proposed dataset. Observe the alternating direction on sequential dials, with the least significant dial (i.e., the rightmost one) always oriented clockwise.}
    \label{fig:illustration-dials}
\end{figure}

The labeling process was a time-consuming task, lasting more than 350 hours in a custom-made tool -- not including the time required to improve the tool itself as well as the relabeling of several images.
During the labeling process, we identified a pattern regarding the direction of the dials (\gls*{cw} or \gls*{ccw}) in all meters in the proposed dataset: the dials have alternating directions, with the least significant dial --~i.e., the rightmost one~-- always oriented clockwise.
Thus, the orientation is always (\gls*{cw}, \gls*{ccw}, \gls*{cw}, \gls*{ccw}, \gls*{cw}) for the 5-dial meters and (\gls*{ccw}, \gls*{cw}, \gls*{ccw}, \gls*{cw}) for the 4-dial meters, as illustrated in Fig~\ref{fig:illustration-dials}.
This information can be useful for training methods focused on the angle of the pointer, regardless of its orientation.
In other words, it is possible to correctly read the dials even in cases where the digits are not legible, that is, considering only the angle of the pointer, since the orientation can then be retrieved using the number of detected dials and their index on the counter~region.

\begin{figure}[!ht]
   \centering
   \includegraphics[width=0.99\linewidth]{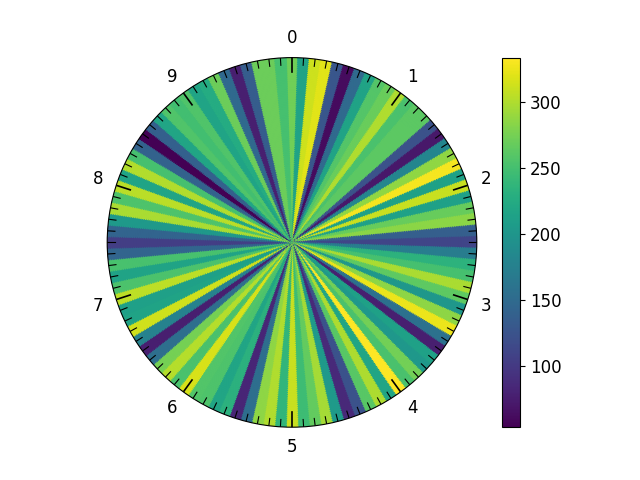}
   
   \vspace{-3.5mm}
   
    \caption{Distribution of the pointer positions in the \dataset dataset.
    }
    \label{fig:dial-distribution}
\end{figure}

Fig.~\ref{fig:dial-distribution} shows the distribution of the dial classes in the dataset, with one decimal place precision.
Note that there is an almost uniform distribution across the digits; however, the values pointed most often are close to the digits themselves and far from the middle position between two~digits.

\subsection{Evaluation Protocol and Metrics}

Following the protocol defined in~\cite{salomon2020deep}, the proposed dataset is divided into three disjoints subsets: $3{,}000$ images for training~($60$\%), $1{,}000$ images for validation~($20$\%), and the remaining $1{,}000$ images for testing~($20$\%).
The subsets are explicitly defined along with the \dataset~dataset, as it is common in  the \gls*{amr} literature~\cite{yang2019,salomon2020deep, laroca2021towards}.
Note that we maintained the original split for the images from the first version of the dataset, \ie, the \ufpradmr's training images are in \dataset's training subset, and the same is true for the validation and test~subsets.

We propose the use of three metrics for evaluation: (i)~\acrfull*{mrr}, (ii)~\gls*{drr}, and (iii)~\acrfull*{mae}.
In the following paragraphs, we detail~them.

The \gls*{mrr} is obtained by comparing, for each of the $N$~meters, the predicted sequence~($pred_m$) with the ground-truth sequence~($gt_m$):
\begin{equation}
    MRR = \frac{1}{N}\sum_{m=1}^N match(pred_m,gt_m) 
\end{equation}

\[
    match(x,y)= 
\begin{cases}
    1,  & \text{if } x = y,\\
    0,  & \text{if } x \neq y.
\end{cases}
\]

The \gls*{drr} is based on the \distance distance (also known as edit distance), a common method for measuring the distance between two sequences of characters. 
This metric has the advantage of dealing with certain minor errors (e.g., an undetected dial, resulting in strings of different lengths) without heavily punishing them, as the distance between the predicted sequence and the ground truth is calculated and divided by the longest sequence. 
Finally, subtracting the error from $1$ results in the dial recognition rate for each meter.
The \gls*{drr} corresponds to the mean of the recognition rates:
\begin{equation}
    DRR = \frac{1}{N} \sum_{m=1}^N \Big( 1 -   \frac{lev_{(pred_m,gt_m)}(|{}pred_m|{},|{}gt_m|{})}{max(|{}pred_m|{}, |{}gt_m|{})}  \Big)
\end{equation}

\gls*{mae} is useful for penalizing errors in the most significant dials (i.e., the leftmost ones).
Considering the sequences as integers, we can obtain the mean absolute error as~follows:

\begin{equation}
    MAE = \frac{1}{N} \sum_{m=1}^N |{}p_m - g_m|{}
\end{equation}
\section{Evaluated Approaches}
\label{sec:approaches}

\begin{figure*}[!htb]
    \centering
    \captionsetup[subfigure]{captionskip=1.5pt}
        
    \subfloat[Detection-based pipeline.\label{fig:detection-pipeline}]{
        \includegraphics[width=0.95\linewidth]{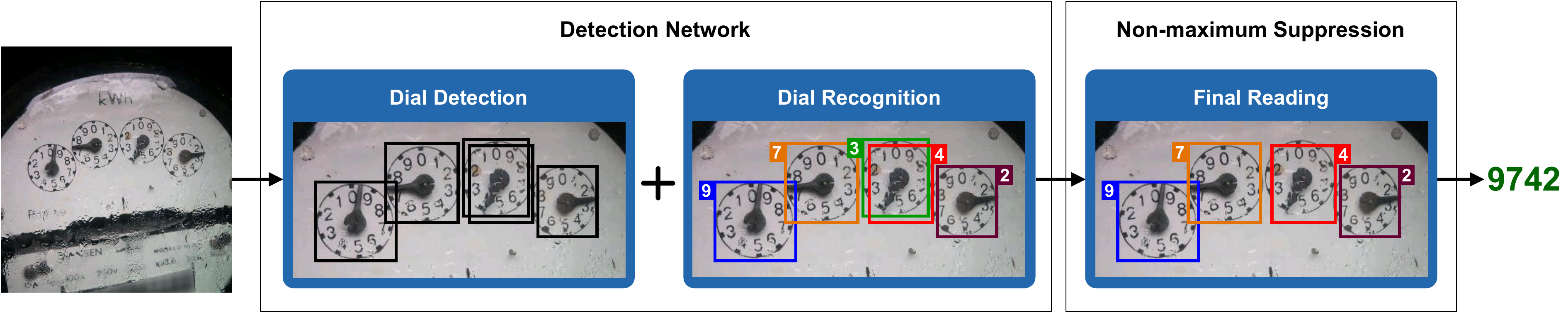}
    }
    
    \vspace{2.25mm}
    
    \subfloat[Regression-based pipeline (AngReg).\label{fig:regression-pipeline}]{
        \includegraphics[width=0.95\linewidth]{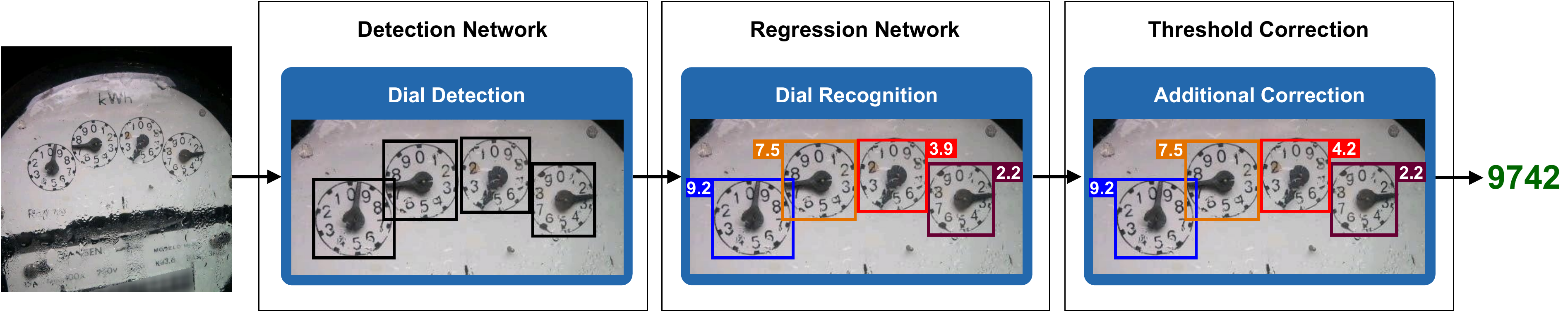}
    }
    
    \vspace{2.25mm}
    
    \subfloat[Segmentation-free pipeline.\label{fig:segmentation-pipeline}]{
        \includegraphics[width=0.95\linewidth]{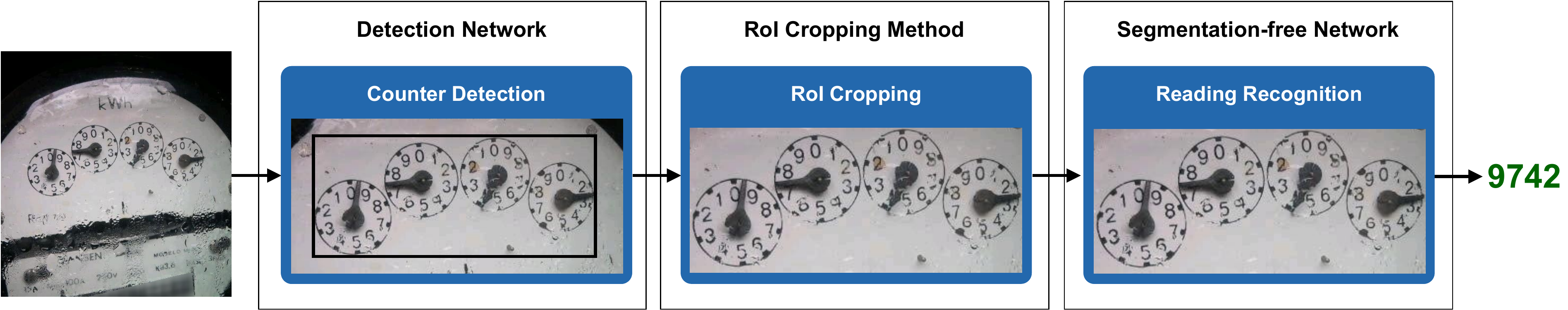}
    }
    
    \vspace{-1mm}
     
    \caption{The proposed pipelines for \gls{admr}. Note that in the detection network the images are zoomed-in for better viewing. In fact, the dial and counter detection are performed on the entire~image.
    }
    \label{fig:pipeline}
\end{figure*}

In this paper, we evaluate three different approaches for \gls*{admr}: (i)~object detection with recognition; (ii)~object detection with regression; and (iii)~segmentation-free recognition. 
The pipeline for each approach is illustrated in Fig~\ref{fig:pipeline}.

\subsection{Object Detection with Recognition}

As can be seen in Fig.~\ref{fig:detection-pipeline}, this approach consists of using object detection networks to simultaneously detect and classify the dials directly in the input image.
The networks are trained to predict $10$ classes (i.e., $0$–$9$) using the bounding box of each dial as input.
Afterward, we apply a \gls*{nms} algorithm to eliminate redundant detections (those with \gls*{iou}~$\ge0.5$) since the
same dial might be detected/recognized more than once by the~network.

For this approach, we decided to evaluate YOLOv3~\cite{redmon2018yolov3} and YOLOv4~\cite{bochkovskiy2020yolov4} considering that YOLO-based models been successfully employed for various computer vision tasks~\cite{kou2021development,laroca2021efficient} and have achieved promising reading results in images of digit-based~meters~\cite{liao2019reading,laroca2021towards}.

Shallower detection models, such as those used by Laroca \etal in~\cite{laroca2019convolutional}~(CR-NET) and in~\cite{laroca2021towards}~(Fast-OCR), also fit this approach. 
However, given the smaller size of these networks, they are fed with the counter region previously detected by another detection network and not directly in the input~image.

\subsection{Object Detection with Regression (AngReg)}
\label{sec:proposed:angreg}

We introduce a novel regression approach, called AngReg, for \gls{admr}.
AngReg comprises a detection model, followed by a custom regression model (with trigonometric activation functions) and a novel post-processing correction~approach.

Similar to the object detection approach, AngReg also leverages an object detector (i.e., YOLOv4~\cite{bochkovskiy2020yolov4}) to obtain the position of all dials from the input image (see Fig~\ref{fig:regression-pipeline}). 

After the detection stage, fine-tuned ResNet~\cite{he2016deep} and Xception~\cite{chollet2017xception} models are explored to extract image features, and a custom \gls*{fcn} with the hyperbolic tangent functions is used to predict the reading for each detected dial.
\major{The AngReg topology is described in more detail in \cref{sec:appendix-angreg}.}

\gls{admr} can be addressed as a regression problem since we can introduce more precision to the pointer prediction (\eg,~$5.75$, while object detection approaches usually yield a discrete label (\eg,~`$5$').
This added precision enables the application of our correction method (described below), which provides information on whether the next dial's prediction matches the current one based on the decimal place (i.e.,~if the decimal place is too far from the integer part of the next dial, probably one of the predictions is not~accurate).

For training the AngReg regression models, we evaluated several target functions such as linear, sigmoid, and hyperbolic tangent.
The labels were converted according to the activation function for the sigmoid and tangent targets.
For the sigmoid function, we scaled the label values between $0.0$ and $1.0$.
We converted the labels values to angles for the tangent function and used the sine and cosine to obtain the polar coordinates correctly.
The encoding of the angles is performed using the following formulation to obtain Cartesian coordinates from polar coordinates (assuming the radius is $1$):
\begin{equation}
   \theta \mapsto (\sin(\theta), \cos(\theta)).
\end{equation}

After this conversion, the target function is composed of two hyperbolic tangent perceptrons (one for the sine coordinate and the other for the cosine coordinate --~the $y$ and $x$ coordinates, respectively).
To predict the correct angle for a cropped dial, we need to convert those two Cartesian coordinates to polar coordinates (in our case, to an angle $\theta$).

To convert the coordinates back from Cartesian to polar we use the following formulation:
\begin{equation}
    (y,x) \mapsto \arctan2(y,x) 
\end{equation}

\begin{equation*}
    \arctan2(y,x) =
\begin{cases}
 \arctan\left(\frac{y}{x}\right) &\text{if } x > 0, \\
 \frac{\pi}{2} - \arctan\left(\frac{x}{y}\right) & \text{if } y > 0, \\
 -\frac{\pi}{2} - \arctan\left(\frac{x}{y}\right) & \text{if } y < 0, \\
 \arctan\left(\frac{y}{x}\right) \pm \pi & \text{if } x < 0, \\
 \text{0} & \text{if } x = 0 \text{ and } y = 0.
\end{cases}
\end{equation*}

After the predicted angle is yield, we can obtain the numerical value for the prediction (ranging between $0.0$ and $9.9$) using the orientation pattern (see Section~\ref{sec:dataset}) to determine whether the dial is clockwise or counterclockwise~oriented.

After the numerical value is retrieved, we apply a heuristic method to correct certain reading errors.
This method uses the dial immediately on the right to check whether the predicted value for a given dial is accurate.
For instance, if the current dial's prediction is~$3.9$, but the next dial's prediction is~$2.2$, it is very likely that the correct reading for the current dial is~$4$ instead of~$3$.
This case is illustrated in Fig.~\ref{fig:regression-pipeline}.
Note that the rightmost dial (the least significant one) cannot be corrected using this method, as there is no dial to the right to imply the correct~value.

We used the validation subset to find the best thresholds for this correction method.
These thresholds are basically two pairs of values (i.e., the lower and upper bounds for a given dial and the one immediately to the right) that indicate whether to correct the prediction for the current dial using the reading value predicted for the dial to the~right. 

Fig.~\ref{fig:regression-pipeline} shows an example where this correction method is applied.
Observe that the dial pointing to $3.9$ is followed by a dial pointing to $2.2$ (these values were predicted by the regression model).
Suppose the upper correction threshold is $N.75$ for the dial being analyzed, and the lower correction threshold for the next dial is $N{.}5$. 
As $3.9 > N.75$ (considering only the decimal values) and $2.2 < 2.5$, we correct the predicted value for the dial being analyzed to $4.2$.
Lastly, the float values are rounded to integers to produce the final~reading.

\subsection{Segmentation-free Recognition}

The segmentation-free approach refers to the meter reading being predicted holistically from the counter region, \ie, deep models are employed to predict the meter reading without the need for detecting/segmenting each of its dials. 
Such models have been successfully explored in \gls*{ocr} applications~\cite{calefati2019reading,atienza2021vitstr,laroca2022cross}.

For this approach, it is necessary to first detect the \gls*{roi} (see Fig.~\ref{fig:segmentation-pipeline}) due to the fact that in unconstrained scenarios the \gls*{roi} can occupy a very small portion of the input image~--~this stage is commonly referred to as ``counter detection'' in the \gls*{amr} literature.
For this task, we used a lightweight detection network~\cite{laroca2021towards}, called \detnet, as it was robust enough to reach $100$\% F-score (\gls*{iou}~$\ge0.5$) in our~experiments.

We then evaluated several segmentation-free models well known in the literature.
Some of them are multi-task networks, \ie, those proposed or adapted by G\'{o}mez et al.~\cite{gomez2018cutting}, Laroca et al.~\cite{laroca2019convolutional} (Multi-task), and Calefati et al.~\cite{calefati2019reading}, while others are recurrent-, sequential- or Transformer-based networks commonly employed for scene text recognition: CRNN~\cite{shi2017endtoend}, GRCNN~\cite{wang2017deep}, \rtwoam~\cite{lee2016recursive}, \rare~\cite{shi2016robust}, \rosetta~\cite{borisyuk2018rosetta}, STAR-Net~\cite{liu2016starnet}, TRBA~\cite{baek2019what}, and ViTSTR-Base~\cite{atienza2021vitstr}.

\subsection{Data Augmentation}

We explored various data augmentation strategies to train the evaluated networks, such as random cropping, conversion to grayscale, and random perturbations of hue, saturation and brightness.
\major{Our main goal with data augmentation is to reduce lighting and symmetry errors (those errors occur due to the lack of sufficient and heterogeneous data). 
Thus, we chose the most common lighting, color and position data augmentation techniques used in the literature~\cite{laroca2019convolutional,yang2022its}.
While there are several data augmentation libraries available, we created the modified images using Albumentations~\cite{albumentations}, which is a well-known Python library for image augmentation that has been successfully used in several applications, including \gls*{amr}~\cite{laroca2021towards}.}

In preliminary experiments, we also evaluated the possibility of ignoring the dial direction, forcing the methods to learn the pointer angle rather than the digit and dial direction.
For this purpose, we changed the labels on the \gls*{ccw} dials to match the \gls*{cw} dials ($1$ became $8$, $2$ became $7$, and so on).
The label can be converted back after prediction using the dial direction pattern described in Section~\ref{sec:dataset}.
However, contrary to what we expected, there was no significant improvement in the results achieved (further investigation is needed in this~regard).

\subsection{Post-correction}
\label{sec:post-correction}

\begin{figure}[!ht]
   \centering
    \includegraphics[width=0.99\linewidth]{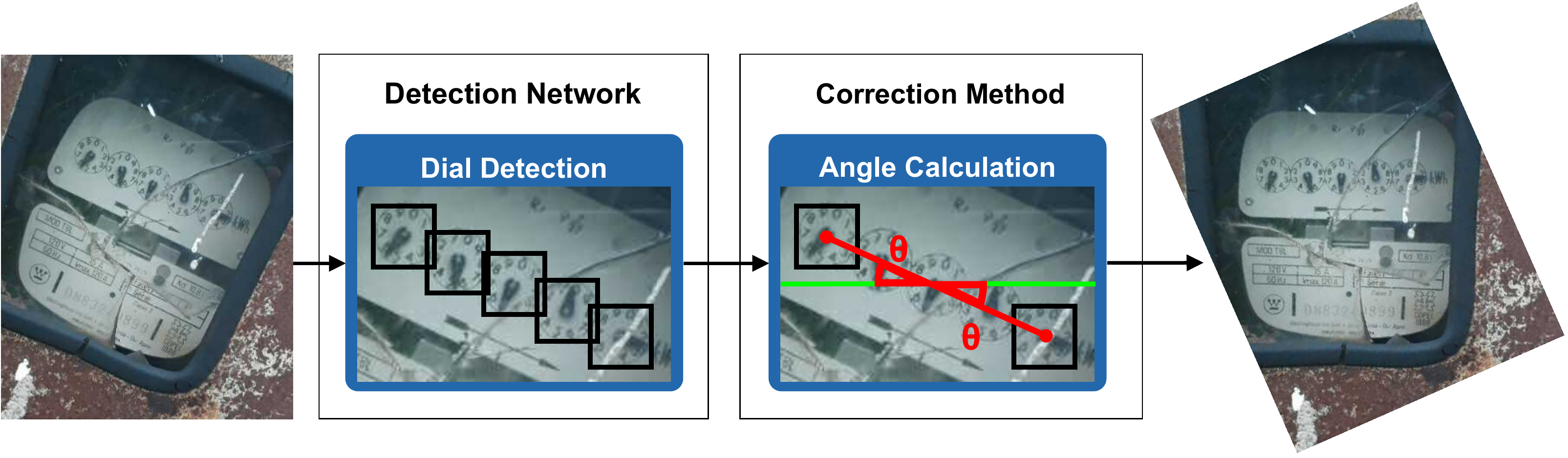}
    
    \vspace{-1.25mm}
    
    \caption{Illustration of the angle correction method. The angle~($\theta$) between the line (in red) that passes through the middle of the first and last detected dials and the $x$-axis (in green) is computed.
    Afterward, the correction method rotates the image -$\theta$ degrees in order to make the dials parallel to the~$x$-axis.}
    \label{fig:correction-pipeline}
\end{figure}

Considering that the images in the \dataset dataset were acquired in unconstrained environments and, therefore, some images have the meter rotated at an angle of almost~$45\degree$, we leverage a post-processing strategy to improve the results obtained by the detection-based approaches.
First, for the predictions obtained for each meter, we calculate the angle of the line segment that passes through the middle of the first and last dial bounding boxes. 
Then, if this angle is greater than a defined threshold (in this work, $\theta>2.5\degree$), we rotate the image accordingly so that the angle with the $x$-axis is approximately~$0\degree$;
\major{note that this threshold ($\theta>2.5\degree$) was defined based on the results achieved in the validation set.
In these experiments, rotating the images in which the angle was less than $2.5\degree$ did not improve the results.}
Lastly, we feed the corrected image to the network and consider these predictions the final reading.
An illustration of the whole process is provided in Fig~\ref{fig:correction-pipeline}.

\section{Experiments}
\label{sec:results}

\major{In this section, we report the experiments performed to verify the effectiveness of the evaluated approaches.
}

\subsection{\major{Setup, Deep Models, and Evaluation Protocol}}

All our experiments were carried out on an AMD Ryzen Threadripper $1920$X $3.5$GHz CPU, $96$~GB of RAM, and an NVIDIA Quadro RTX $8000$ GPU.
For a fair comparison, we trained all approaches (detection, regression, and segmentation-free) using exactly the same~images.

\major{Table~\ref{tab:models} presents an overview of the $\nummodels$ deep models we trained and tested for this work (these models were mentioned throughout \cref{sec:approaches}), listing the framework we used to implement each of them.
These models were chosen mainly because they have been explored for several tasks with promising/impressive results~\cite{calefati2019reading,gomez2018cutting,atienza2021vitstr} and due to the fact that we believe we have the necessary knowledge to train/adjust them in the best possible way in order to ensure fairness in our experiments, as the respective authors provided enough details about the architectures used and because we designed/used similar networks in previous works~\cite{goncalves2019multitask,salomon2020deep,laroca2021towards}.}

\major{
Each model was trained and tested using either the framework where it was originally implemented or well-known public repositories.
In summary, the YOLO-based models (detection networks) were trained using the Darknet framework\footnote{\url{https://github.com/AlexeyAB/darknet/}}, the multi-task models were trained using Keras\footnote{\url{https://keras.io/}}, and the other models were trained using a fork of the open source repository of Clova AI Research\footnote{\url{https://github.com/roatienza/deep-text-recognition-benchmark/}}, which was employed to record the 1st place of ICDAR2013 focused scene text and ICDAR2019~ArT, and 3rd place of ICDAR2017COCO-Text and ICDAR2019 ReCTS~(task1).
We provide more details on the implementation of these methods in \cref{sec:appendix-hyperparameters}.
}

\begin{table}[!htb]
\centering
\caption{\major{Deep models explored in our experiments along with the respective frameworks used to implement them.}}
\label{tab:models}

\vspace{-1.5mm}

\major{\begin{tabular}{@{}lc@{}}
\toprule
\multicolumn{1}{c}{Model} & Framework                          \\ \midrule
\hspace{3mm}CR-NET~\cite{silva2020realtime} & Darknet \\   
\hspace{3mm}Laroca et al.~\cite{laroca2021towards} & Darknet \\
\hspace{3mm}YOLOv3~\cite{redmon2018yolov3} & Darknet \\
\hspace{3mm}YOLOv4~\cite{bochkovskiy2020yolov4} & Darknet \\ \midrule \midrule
\hspace{3mm}Calefati et al.~\cite{calefati2019reading} & Keras \\ 
\hspace{3mm}G\'{o}mez et al.~\cite{gomez2018cutting} & Keras \\ 
\hspace{3mm}\multitaskgabriel~\cite{goncalves2019multitask} & Keras \\  
\hspace{3mm}ResNet50~\cite{he2016deep} & Keras \\
\hspace{3mm}ResNet101~\cite{he2016deep} & Keras \\ 
\hspace{3mm}Xception~\cite{chollet2017xception} & Keras \\ \midrule \midrule
\hspace{3mm}\crnn~\cite{shi2017endtoend} & PyTorch                                \\
\hspace{3mm}\grcnn~\cite{wang2017deep} & PyTorch                                \\
\hspace{3mm}\rtwoam~\cite{lee2016recursive}  & PyTorch                                  \\
\hspace{3mm}\rare~\cite{shi2016robust}   & PyTorch                             \\
\hspace{3mm}\rosetta~\cite{borisyuk2018rosetta}  & PyTorch                           \\
\hspace{3mm}\starnet~\cite{liu2016starnet}  & PyTorch                           \\
\hspace{3mm}\trba~\cite{baek2019what}  & PyTorch                                \\
\hspace{3mm}\vitstrbase~\cite{atienza2021vitstr}   & PyTorch                       \\
\bottomrule
\end{tabular}%
}
\end{table}

For the detection networks, we used the \emph{\gls*{map}} evaluation metric to stop the training process and select the best model for testing, as it has been commonly employed on object detection tasks~\cite{ren2015faster,redmon2018yolov3,bochkovskiy2020yolov4}. 
The \gls*{map} can be calculated as~follows:
\begin{equation}
    mAP = \frac{1}{c} \sum_{i=1}^c{AP_i} \, ,
\end{equation}
\noindent where $AP_i$ stands for the average precision value (for recall values from $0$ to $1$) of the $i$-th~class. 

We evaluated several metrics such as \gls*{rmse}, \gls*{mse}, and \gls*{mae} for the regression approaches.
The best results were achieved using the \gls*{mse}, which can be obtained by:
\begin{equation}
   MSE = \frac{1}{n}\sum_{t=1}^{n}e_t^2
\end{equation}

\subsection{\major{Results}}

The results obtained by all approaches and models evaluated are shown in Table~\ref{tab:results-recognition}.
In general, the detection-based models (e.g., YOLOv3 and YOLOv4) outperformed both regression and segmentation-free models for the \gls*{admr} task.
Usually, complete systems (which perform detection and recognition using the same network) perform better than the sum of its isolated parts~\cite{goncalves2019multitask, laroca2021efficient}.
This is due to the fact that two-stage methods usually propagate the error from earlier stages through the entire pipeline.
In addition, minimizing each loss separately is solving two different tasks, instead of jointly minimizing both losses, which is YOLO's target. Furthermore, segmentation-free approaches usually require a higher amount of data to train, as seen in~\cite{gomez2018cutting,laroca2019convolutional}.

\begin{table}[!htb]
 \centering
 
 \caption{Reading results achieved on the \dataset dataset.}
 
 \vspace{-1.5mm}
 \resizebox{0.99\linewidth}{!}{
 \begin{tabular}{lccr}
 \toprule
 \multirow{2}{*}{Approach} & \multicolumn{2}{c}{Recog. Rate (\%)} & \multirow{2}{*}{MAE} \\
 
  & Meter & Dial &  \\ 
 \midrule

Laroca et al.~\cite{laroca2019convolutional} (Multi-task~\cite{goncalves2019multitask}) & $15.30$ & $64.24$ & $24{,}181$ \\
Calefati et al.~\cite{calefati2019reading} & $21.00$ & $72.20$ & $20{,}078$ \\ 
G\'{o}mez et al.~\cite{gomez2018cutting} & $23.50$ & $70.56$ & $24{,}074$ \\ 
\rosetta~\cite{borisyuk2018rosetta} & $57.50$ & $82.39$ & $4{,}968$ \\ 
\rtwoam~\cite{lee2016recursive} & $59.60$ & $83.72$ & $4{,}087$ \\ 
STAR-Net~\cite{liu2016starnet} & $60.00$ & $83.70$ & $9{,}454$ \\ 
GRCNN~\cite{wang2017deep} & $60.80$ & $84.14$ & $6{,}108$ \\ 
\rare~\cite{shi2016robust} & $61.80$ & $84.25$ & $5{,}696$ \\ 
CRNN~\cite{shi2017endtoend} & $64.20$ & $86.27$ & $3{,}669$ \\ 
TRBA~\cite{baek2019what} & $66.80$ & $87.87$ & $2{,}404$ \\ 
Laroca et al.~\cite{laroca2019convolutional} (CR-NET~\cite{silva2020realtime}) & $71.40$ & $91.94$ & $1{,}680$ \\ 
\vitstrbase~\cite{atienza2021vitstr} & $76.20$ & $92.45$ & $2{,}841$ \\ 
Laroca et al.~\cite{laroca2021towards} & $78.30$ & $94.07$ & $1{,}327$ \\

AngReg (ResNet50-based~\cite{he2016deep}) & $85.00$ & $96.38$ & $429$ \\
AngReg (ResNet101-based~\cite{he2016deep}) & $86.20$ & $96.52$ & $402$ \\
AngReg (Xception-based~\cite{chollet2017xception}) & $88.00$ & $96.96$ & $244$ \\

YOLOv3~\cite{redmon2018yolov3} & $88.60$ & $97.00$ & $477$ \\

YOLOv4~\cite{bochkovskiy2020yolov4} & $91.10$ & $97.67$ & $242$ \\ 

YOLOv4~\cite{bochkovskiy2020yolov4} (R) & $92.00$ & $98.00$ & $197$ \\

YOLOv4~\cite{bochkovskiy2020yolov4} (R) + AngReg (Xception-based~\cite{chollet2017xception}) & $\textbf{92.50}$\textbf{} & $\textbf{98.11}$\textbf{} & $\textbf{187}$ \\

 \bottomrule
 \end{tabular}
 }
\label{tab:results-recognition}
\end{table}

\begin{figure*}[!htb]
    \centering

    \includegraphics[width=0.95\linewidth]{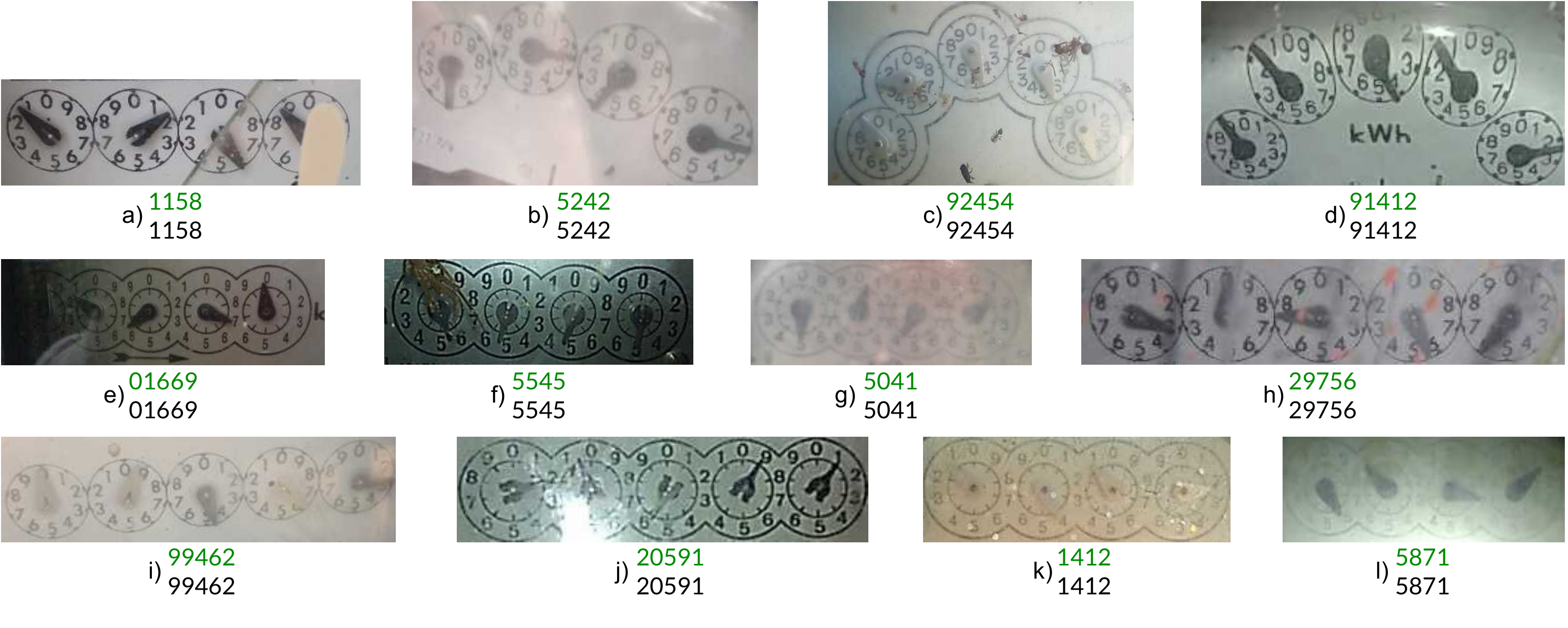}

    \vspace{-6mm}

    \caption{Examples of correct readings performed by the approach that yielded the best results in our experiments: YOLOv4 (R) + AngReg~(Xception). 
    }
    \label{fig:results-hits}
\end{figure*}

The best results were achieved by combining the best object detection model (YOLOv4~\cite{bochkovskiy2020yolov4}) with the best regression model (AngReg (Xception-based~\cite{chollet2017xception})).
This generated our correction strategy, where AngReg indicated when the corrections should be performed, reducing the errors in the rightmost dials.
We used the regression results to verify whether the dial needs to be corrected or not. 
It is noteworthy that this correction approach is only possible due to the approximate floating point annotated labels available in the \dataset dataset.
With an error tolerance of $1$ \gls*{kwh}, our final system achieves an impressive \gls*{mrr} of $98.9$\%.
In the next subsection, we detail why this error tolerance is acceptable.
Samples of correctly predicted readings are shown in~Fig.~\ref{fig:results-hits}. 

These results are significantly better than those achieved in our previous work~\cite{salomon2020deep}.
The reasons are threefold: (i)~there are now more than twice as many images for training the deep networks, which further highlights the importance of the \dataset dataset; (ii)~we fine-tuned several recent approaches and models for \gls*{admr} (as far as we know, there is no other work related to this problem with so many methods being evaluated); and (iii)~we dealt specifically with each of the main causes of reading errors found in~\cite{salomon2020deep}: neighboring values, rotated images, lighting, and mirrored~predictions, as described in the following~subsection.

\subsubsection{Error Analysis}

The main sources of errors in~\cite{salomon2020deep} were caused by:
\begin{itemize}
    \item \textbf{Symmetry:} as there are clockwise and counterclockwise dials, in some cases the recognition methods can not differentiate the direction and thus may output the mirrored value of the actual prediction.
    \item \textbf{Neighboring value:} the most common error. Variables such as angle, lighting, shadows and occlusion can impair the reading of a dial --~especially when the dial is really close to the scale mark (almost changing from one digit to the next one). In addition, mechanical failures can lead to misaligned pointers causing \emph{misalignment} errors (a subtype of neighboring value errors).
    \item \textbf{Severe lighting conditions \& Dirt:} shadows, glares, reflections and dirt may confuse the networks, especially in low-contrast images, where those artifacts may emerge more than the pointer, fooling the network to think that it is a pointer border, resulting in an incorrect prediction. 
    \item \textbf{Rotation:} tilted images are harder to predict, as the pointed value is not in the usual position. The predictions may be assigned to neighboring digits that would be in the current angle of the pointer if the image was not~tilted.
\end{itemize}

The neighboring values issue was tackled by combining the regression with the detection approach. The regression approach uses the neighboring value to correctly yield a prediction, thus achieving a lower \gls*{mae} even though the \gls*{mrr} and \gls*{drr} are slightly worse than~YOLOv3.

Symmetry- and lighting-related errors were addressed using various data augmentation techniques to create many images with different characteristics from a single labeled one.
There was only one symmetry error made by the best model (i.e., YOLOv4 (R) + AngReg(Xception)).

As mentioned in Section~\ref{sec:post-correction}, we solved the rotation issue using the angle between the first and last dial bounding boxes to rotate and rectify the dial image. 
This strategy improved the recognition rate achieved by YOLOv4 by~$0.9$\%.

\begin{figure}[!htb]
   \centering
    \includegraphics[width=0.85\linewidth]{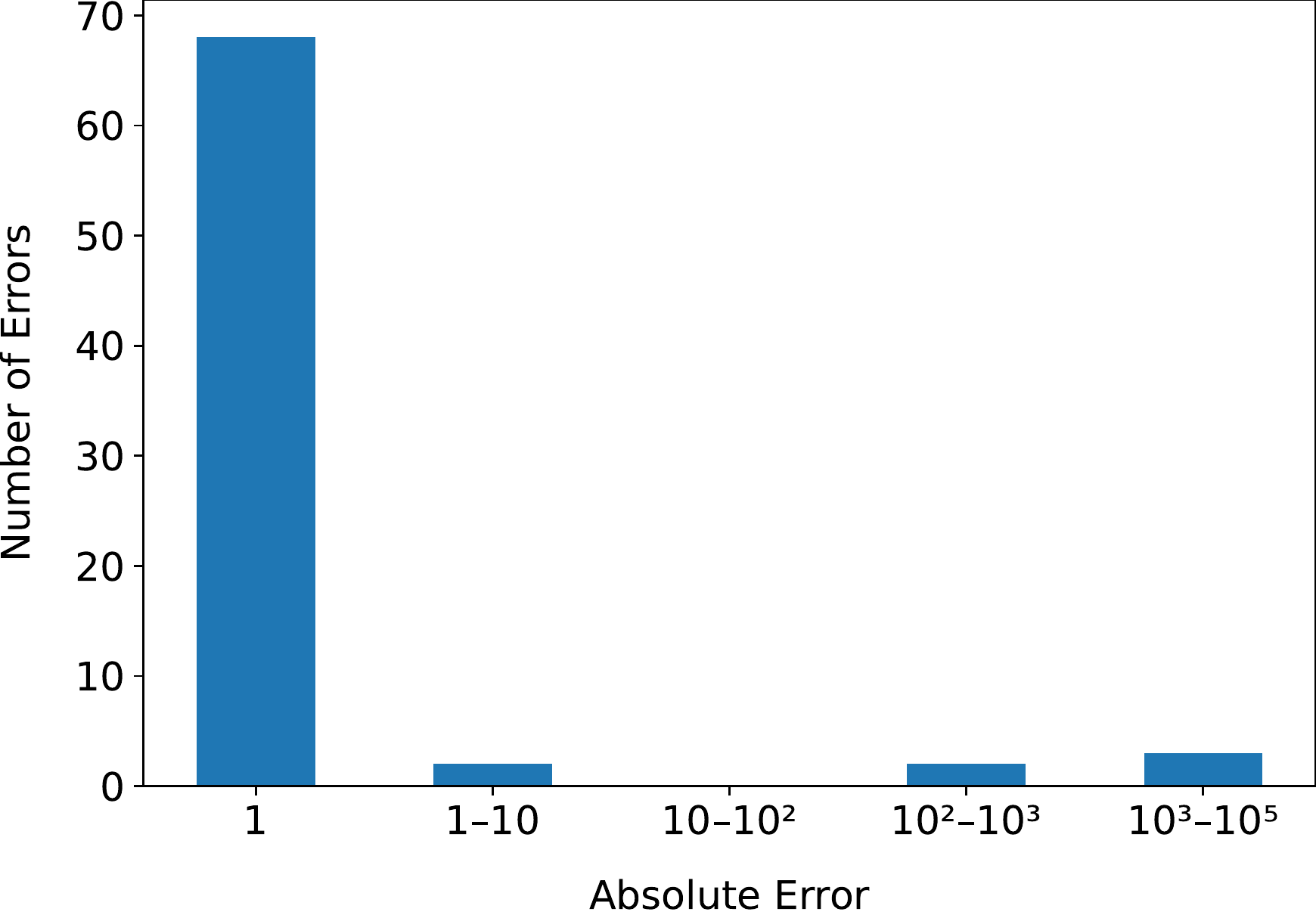}
    
    \vspace{-1.5mm}
    
    \caption{Illustration of the number of errors per absolute error range. Note that most of the errors ($85$\%) have MAE~$=$~$1$. The number of errors refers to the $1{,}000$ images from the test set of \dataset.}
    \label{fig:error-magnitude}
\end{figure}

\begin{figure*}[!htb]
    \centering

    \includegraphics[width=0.95\linewidth]{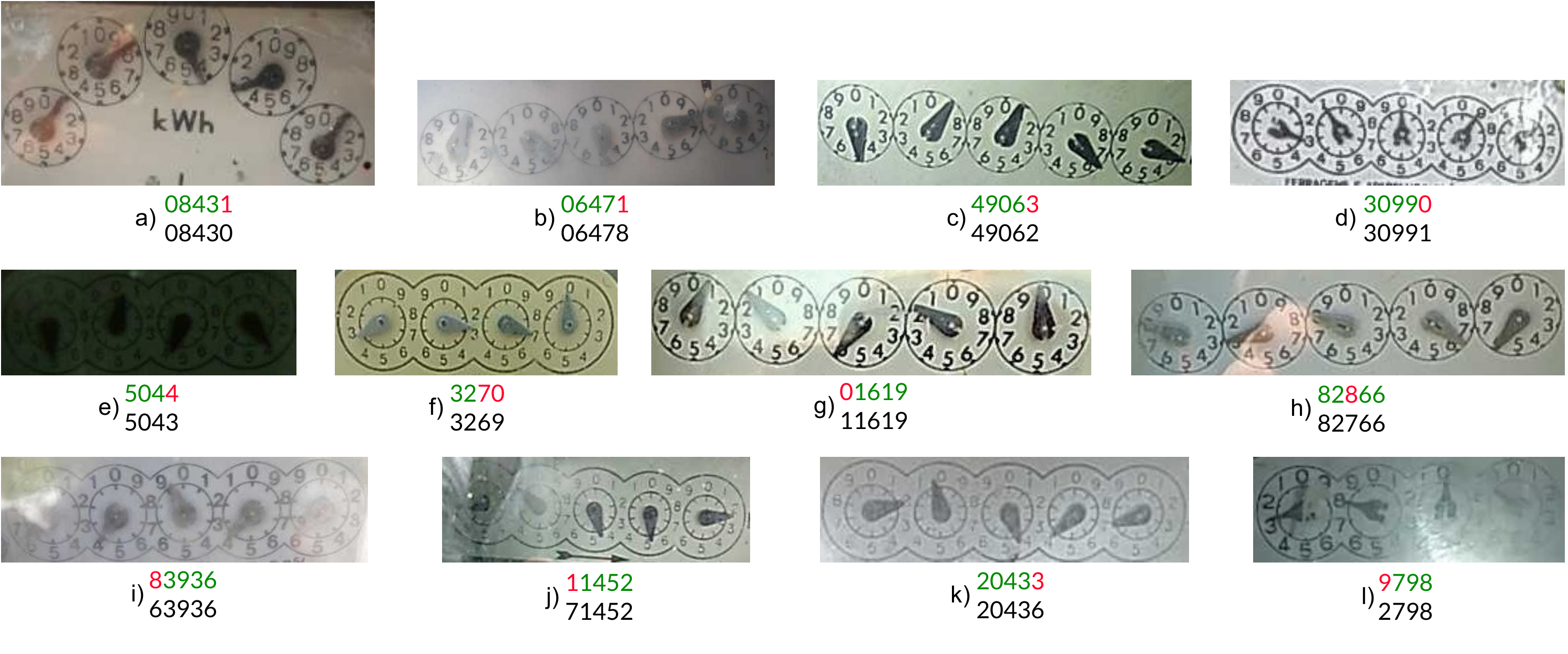}

    \vspace{-6mm}
    
    \caption{
    Examples of reading errors made by the approach that achieved the best results in our experiments: YOLOv4 (R) + AngReg~(Xception).
    The prediction (top reading) and the ground truth (bottom reading) are listed for each meter. 
    The reading errors are outlined in red for better viewing. The error types are neighboring values~(a, c, e, f), misalignment~(g, h), severe lighting conditions~(d, e, i, j), symmetry~(k), and dirt~(b, l).
    }
    \label{fig:results-errors}
\end{figure*}

As listed in Table~\ref{tab:results-dist-errors}, most errors occurred on the least significant dial (i.e., the rightmost one). 
Although there are multi-dial errors, as shown in Fig.~\ref{fig:results-errors}, these errors have \gls*{mae}~$=1$.
In fact, most of the errors ($85$\%) have \gls*{mae}~$=1$, as shown in Fig.~\ref{fig:error-magnitude}.
We estimate that such an error would only cost approximately R\$~$0.83$ (US\$~$0.16$) to the customer\footnote{\url{https://www.copel.com/hpcweb/copel-distribuicao/taxas-tarifas/}} and would probably be compensated in the next monthly~reading.

\begin{table}[!htb]
 \centering
 \caption{Distribution of errors by dial position.}
 
 \vspace{-1.5mm}
 \resizebox{.99\linewidth}{!}{
 \begin{tabular}{lccccc}
 \toprule
 \multirow{2}{*}{\diagbox[trim=l,trim=r,innerrightsep=-2.5pt,font=\footnotesize]{Approach}{Dial Position}} & \multicolumn{5}{c}{Frequency (\%)}   \\
 & $1$ & $2$ & $3$ & $4$ & $5$ \\
\midrule

 YOLOv3~\cite{liao2019reading} & $7.86$ & $8.57$ & $7.14$ & $11.43$ & $65.00$  \\
 YOLOv4~\cite{bochkovskiy2020yolov4}  & $8.49$ & $5.66$ & $5.66$ & $15.09$ & $65.09$ \\
 YOLOv4~\cite{bochkovskiy2020yolov4} (R) & $5.21$ & $3.12$ & $6.25$ & $12.50$ & $72.92$ \\
 YOLOv4~\cite{bochkovskiy2020yolov4} (R) + AngReg (Xception-based~\cite{chollet2017xception}) & \textbf{$4.71$} & $1.18$ & $3.53$ & \textbf{$10.59$} & \textbf{$80.00$} \\

 \bottomrule

 \end{tabular}
 }
\label{tab:results-dist-errors}
\end{table}

Moreover, it is hard even for humans to identify the correct reading due to perspective, lighting, shadows, and image quality.
Indeed, there were disagreements between the authors regarding the correct label in some cases, which makes this type of error really hard to avoid without a policy of rounding up or down the reading value in such~scenarios. 

As previously mentioned, with an error tolerance of $1$ \gls*{kwh}, our final system achieves an impressive \gls*{mrr} of $98.90$\%.
Thus, we consider that the YOLOv4~(R) + AngReg~(X) model can be reliably employed on real-world meter reading applications. 
However, there are still opportunities for enhancements since most of the errors occurred on images with inadequate quality and lighting, which should be double-checked by a reading specialist or rejected in a real-world~system.

Finally, the service company could easily detect the most significant errors using simple heuristic rules or even through a model capable of predicting the consumption of each consumer based on previous readings.
If the reading is lower than the one recorded in the previous month or indicated that the consumer suddenly used a huge amount of energy, it can be double-checked by~employees.

\section{Conclusions}
\label{sec:conclusions}

We tackled most of the issues regarding the automatic reading of dial meters.
We came up with an approach that combines a detection model with a proposed new regression approach (AngReg) to enable error correction for the most significant dials, thus decreasing the \gls*{mae} considerably.
We believe that such a robust approach can help reduce meter reading costs, reading errors, tampering and theft of meters, thus generating substantial savings for service~companies.

\major{More specifically, the neighboring values error -- the most frequent type of error in~\cite{salomon2020deep} -- was tackled using AngReg.
Errors related to symmetry and lighting were addressed by creating many new training images through data augmentation.
Lastly, we eliminated rotation errors by rotating the images according to the calculated angle between the detected dials.
}

We also introduced the \dataset dataset, which is a public dataset for \gls*{admr} that includes $5{,}000$ fully-annotated meter images acquired in unconstrained scenarios by the service company’s employees themselves.
Compared to the \ufpradmr dataset, introduced in our previous work~\cite{salomon2020deep}, the expanded version has more than twice the images and contains new annotations such as the approximated pointed value labels (with one decimal place precision) and segmentation masks for the~dials. 
Such annotations enable the application of a much wider range of techniques for~\gls*{admr}.

We also provided a detailed error analysis in order to assist the deployment of real-world \gls{admr} systems.
We showed that, by accepting minor errors (i.e., with \gls*{mae}~=~$1$) that have negligible cost (less than US\$$0.16$), we could increase the system's \gls{mrr} from $92.50$\% to $98.90$\%.
In fact, this cost would almost certainly be compensated in a subsequent~reading.

As future work, we intend to develop a rejection system in order to automatically request new image samples in cases of poorly taken images.
Additionally, we plan to leverage \glspl*{gan} to generate more diverse training samples, ideally with pointers pointing equally at all angles (in terms of the number of dials across the images), to reduce bias by improving the balance between~classes.
\section*{Acknowledgments}
\label{sec:acknowledgments}

This work was supported in part by the National Council for Scientific and Technological Development~(CNPq) (Grant~308879/2020-1), and in part by the Coordination for the Improvement of Higher Education Personnel~(CAPES) (Grant 88887.516264/2020-00 and Social Demand Program).
The Quadro RTX $8000$ used for this research was donated by the NVIDIA Corporation.
We also used another GPU (Titan V) donated by the NVIDIA Corporation through Professor Andr\'e Gregio's request.
We gratefully acknowledge the \acrfull*{copel}, particularly the manager of the reading division Dihon Pereira Brand\~{a}o, for providing the images for the creation of the \dataset~dataset.

\balance

\bibliographystyle{IEEEtran}
\typeout{}
\bibliography{bib/bibtex}

\clearpage
\nobalance
\appendix
\subsection{AngReg Topology}
\label{sec:appendix-angreg}
\begin{figure}[!htb]
   \centering
    \includegraphics[width=0.52\linewidth]{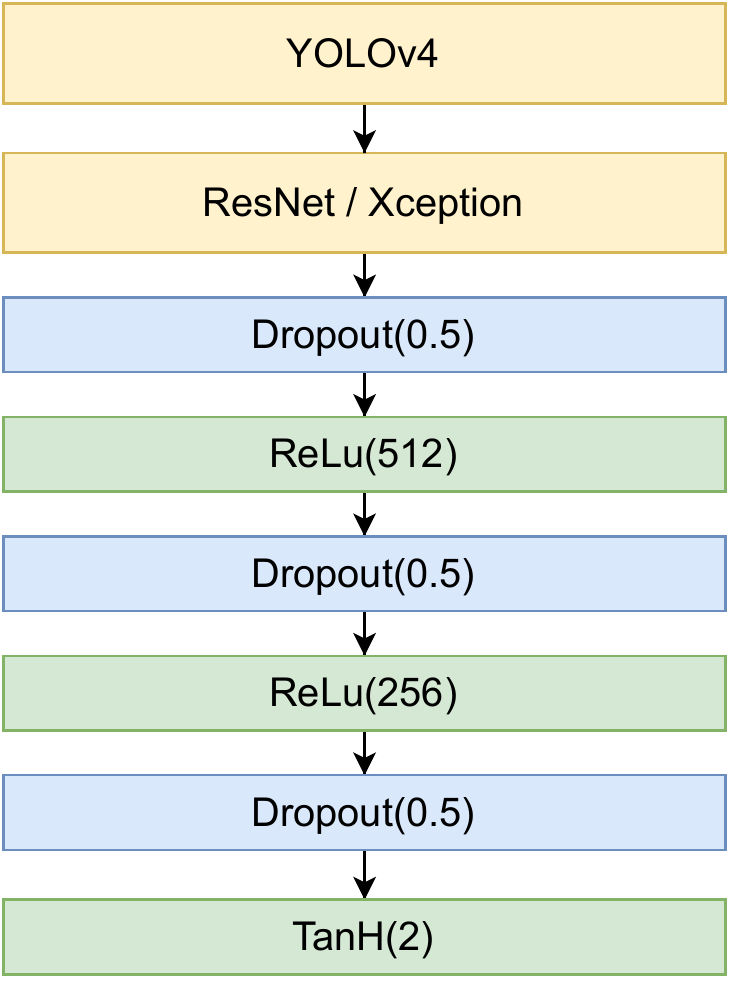}
    
    \vspace{-1.5mm}
    
    \caption{\major{Illustration of the AngReg topology. This architecture was obtained empirically by performing a grid search and optimizing hyperparameters.}}
    \label{fig:angreg-topology}
\end{figure}

\major{
As can be inferred from \cref{sec:proposed:angreg} and Table~\ref{tab:models}, we implemented our AngReg approach using Darknet (YOLOv4) and Keras (ResNet/Xception).
More specifically, as detailed in Fig.~\ref{fig:angreg-topology}, we explored YOLOv4 to first locate all dials in each input image.
Then, we employed the official Keras implementations of ResNet/Xception as an image feature extractor (without the top section).
Finally, a custom \gls*{fcn} was implemented as the classifier, with \gls{mse} as the loss~function.
}

\subsection{Hyperparameters}
\label{sec:appendix-hyperparameters}
\major{
In Darknet, we employed the same parameters as Laroca et al.~\cite{laroca2021towards}:~\gls*{sgd} optimizer, $65$K iterations, batch size~=~$64$, and learning rate~=~[$10$\textsuperscript{-$3$},~$10$\textsuperscript{-$4$},~$10$\textsuperscript{-$5$}] with decay steps at $26$K and $45.5$K.
In Keras, we used the Adam optimizer, learning rate~=~$10$\textsuperscript{-$5$}, batch size~=~$16$-$64$ (depending on the model), max epochs~=~$100$, and patience~=~$15$ (patience refers to the number of epochs with no improvement after which training is stopped); these parameters were defined based on previous works~\cite{calefati2019reading,gomez2018cutting} and also on preliminary experiments in the validation set.
In PyTorch, as in~\cite{baek2019what,atienza2021vitstr}, we adopted the following parameters: Adadelta optimizer (decay rate $\rho=0.99$),  $300$K iterations, and batch size~=~$128$.
}

\end{document}